%% file: log_2024.tex
\documentclass{article}
\usepackage{log_2024}						

\usepackage{booktabs}						
\usepackage{multirow}						
\usepackage{amsfonts}						
\usepackage{graphicx}						
\usepackage{duckuments}						
\usepackage{subcaption}
\usepackage{caption}
\usepackage{wrapfig}
\usepackage{amsmath}
\usepackage{algorithmic}
\usepackage{textcomp}
\usepackage{xcolor}
\usepackage[normalem]{ulem}
\useunder{\uline}{\ul}{}

\usepackage{xspace}
\usepackage{url}
\usepackage{bm}
\usepackage{algorithm}
\usepackage{xcolor}
\newcommand{\tickYes}{{{\checkmark}}}
\usepackage{pifont}
\newcommand{\tickNo}{{\hspace{1pt}{\ding{55}}}}

\usepackage{listings}
\input{dfn}

\newcommand{\method}{{GODM}\xspace}
\newtheorem{definition}{Definition}

\usepackage[numbers,compress,sort]{natbib}	

\title[Data Augmentation for Supervised Graph Outlier Detection via Latent Diffusion Models]{Data Augmentation for Supervised Graph Outlier Detection via Latent Diffusion Models}

\author[K. Liu et al.]{%
Kay Liu$^\dag$ \quad Hengrui Zhang$^\dag$ \quad Ziqing Hu$^\S$ \quad Fangxin Wang$^\dag$ \quad Philip S. Yu$^\dag$\\
$^\dag$University of Illinois Chicago \quad $^\S$University of Notre Dame\\
\email{\{zliu234, hzhan55, fwang51, psyu\}@uic.edu} \quad \email{zhu4@nd.edu}
}

\begin{document}

\maketitle

\begin{abstract}
A fundamental challenge confronting supervised graph outlier detection algorithms is the prevalent problem of \textit{class imbalance}, where the scarcity of outlier instances compared to normal instances often results in suboptimal performance. 
Recently, generative models, especially diffusion models, have demonstrated their efficacy in synthesizing high-fidelity images.
Despite their extraordinary generation quality, their potential in data augmentation for supervised graph outlier detection remains largely underexplored. To bridge this gap, we introduce \method, a novel data augmentation for mitigating class imbalance in supervised \textbf{G}raph \textbf{O}utlier detection via latent \textbf{D}iffusion \textbf{M}odels.
Extensive experiments conducted on multiple datasets substantiate the effectiveness and efficiency of \method. The case study further demonstrated the generation quality of our synthetic data.
To foster accessibility and reproducibility, we encapsulate \method into a plug-and-play package and release it at PyPI: \url{https://pypi.org/project/godm/}.
\end{abstract}

\section{Introduction}
\label{sec:intro}

Graph outlier detection has emerged as a popular and important area of research and practice in graph machine learning~\cite{liu2024pygod, liu2022bond, tang2023gadbench}. Graph outlier detection focuses on detecting outliers within graph-structured data that significantly deviate from standard patterns. It has been proven valuable in domains such as fraud detection~\cite{huang2022dgraph}, fake news detection~\cite{dou2021user}, spam detection~\cite{dou2020robust}, anti-money laundering \cite{weber2019anti}, etc.
Despite the advancements in graph outlier detection techniques, similar to supervised outlier detection on other data modalities, graph outlier detection suffers a fundamental challenge known as class imbalance. This challenge manifests as a significantly lower number of positive instances (outliers) compared to negative instances (inliers). For example, the ratio of positive to negative is only 1:85 on DGraph dataset~\cite{huang2022dgraph}, reflecting the extreme ratio in real-world financial fraud detection scenarios. This class imbalance problem poses challenges in the training of graph outlier detectors and often results in suboptimal performance. Specifically, as the negative instances dominate the training data, the loss function can be biased towards the majority of the negative class and, hence, exhibit poor generalization capability in identifying true outliers. 

\begin{wrapfigure}{r}{0.48\textwidth}
\vspace{-1em}
  \includegraphics[width=0.48\textwidth]{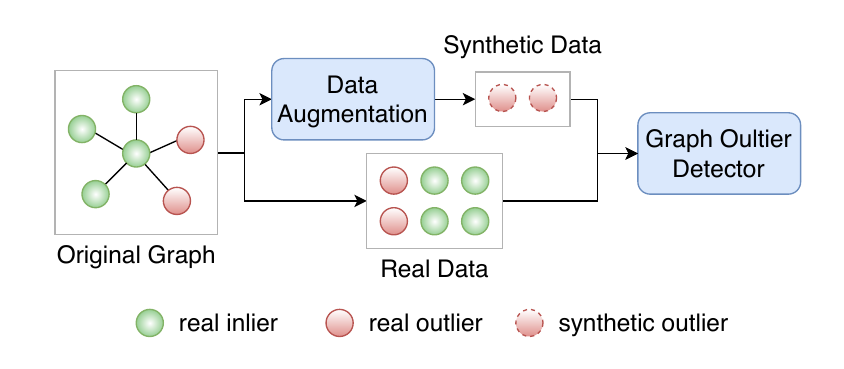}
  \caption{An toy example of data augmentation for class imbalance in graph outlier detection.}
  \vspace{-0.1in}
\label{fig:da}
\end{wrapfigure}

Common practices mitigate this imbalance by upsampling or downsampling \cite{dou2020enhancing}. Upsampling augments the minority positive class by replicating outliers, whereas downsampling reduces the size of the majority negative class by randomly dropping normal instances. However, these methods often have their own risks, such as overfitting outliers in upsampling or the loss of valuable training data through downsampling. Another common approach to alleviate class imbalance is through instance reweighting in the loss function, assigning greater weights to outliers and less weights to inliers. This approach is mathematically equivalent to upsampling and downsampling, thereby having a similar problem. For a more comprehensive summary of previous methods, please refer to Appendix \ref{appx:rw}. These challenges underscore the need for more sophisticated data augmentation methods that can generate some synthetic data to balance the class distribution in the training data, thereby improving the performance of graph outlier detection, as Figure \ref{fig:da} shows.

In recent years, generative models, particularly diffusion models, have achieved significant advancements in synthesizing high-fidelity image data \cite{rombach2022high}. Diffusion models capture intricate data distributions and generate high-quality samples by gradually denoising the samples from a simple prior distribution (e.g., Gaussian distribution). Diffusion models provide more generative power compared to variational autoencoders and offer more stable training than generative adversarial networks. On graph data, although some works have explored the possibility of diffusion on molecular graphs~\cite{jo2022score, vignac2022digress}, few studies have been conducted to apply diffusion models to tackle the class imbalance problem in the task of graph outlier detection. However, existing diffusion models can hardly be directly adapted to large-scale graph outlier detection applications due to the following non-trivial challenges: 
(1) \textbf{Heterogeneity}: Unlike relatively simple molecular graphs, the information contained in graphs for outlier detection can be highly heterogeneous, including high dimensional features and more than one type of edge, even temporal information. Existing diffusion models are primarily designed for monotypic information, exemplified by the RGB channels in image data.
(2)~\textbf{Efficiency}: In the application of graph outlier detection, e.g., financial fraud detection, the graphs are typically much larger than molecular graphs, up to millions, even billions scale. The prohibitive computational cost of diffusion models hinders their direct application in outlier detection on large graphs.
(3) \textbf{Condition}: As we only want to generate outliers rather than normal nodes in synthetic graphs, conditional generation is required to mitigate the problem of class imbalance.

To bridge this gap, we apply diffusion models in graph outlier detection and propose a data augmentation for mitigating class imbalance in supervised \textbf{G}raph \textbf{O}utlier detection via latent \textbf{D}iffusion \textbf{M}odels (\method). Our main idea is to generate outliers in graph space while conducting diffusion in latent space. 
To address heterogeneity, we propose \textit{Variational Encoder} to map the heterogeneous information inherent within the graph data into a unified latent space. In addition, \textit{Graph Generator} synthesizes different types of information back to graph space from the latent embedding. 
To alleviate the efficiency problem, instead of direct diffusion in graph space, we only conduct diffusion in the latent space crafted by the variational encoder. Furthermore, we use negative sampling and graph clustering to reduce the computational cost. For diffusion models, we also adopt EDM \cite{karras2022elucidating} instead of commonly used DDPM \cite{ho2020denoising} to facilitate the generation efficiency of \textit{Latent Diffusion Model}.
For the condition, we not only give a class label to the variational encoder to form node embedding with class information but also conduct conditional generation on both Latent Diffusion Model and Graph Generator. Finally, our heterogeneous, efficient, and conditional \method can generate graphs with outliers that are integrated with the original real graph for the training of the downstream graph outlier detector. Importantly, \method is model agnostic, providing researchers and practitioners with the flexibility to integrate it across various graph outlier detectors. To foster accessibility, we make our code a plug-and-play package, which can be easily adopted on PyG \texttt{Data} object.

\section{Preliminary}
\label{sec:prelim}

In this section, we establish the notation adopted in the subsequent sections and rigorously formulate the problem of data augmentation for addressing class imbalance in supervised graph outlier detection. 

\subsection{Notation}

Let $\mathcal{G} = ( \mathcal{V}, \mathcal{E}, \mathbf{X}, \mathbf{y}, \mathbf{t}, \mathbf{p})$ denotes a graph with $n$ nodes, where $\mathcal{V}=\{ v_i\}_{i=1}^n$ is the set of nodes, and $\mathcal{E}=\{e_{ij}\}$ represents the set of edges. Here, $e_{ij}=(v_i, v_j)$ denotes an edge between node $v_i$ and node $v_j$. The matrix $\mathbf{X} \in \mathbb{R}^{n \times d}$ contains $d$-dimensional feature vectors $\mathbf{x}_i$ for each node $v_i$. \ $\mathbf{y} \in \{0,1\}^n$ represents the vector of the node label $y_i$ for each node $v_i$, where $0$ denotes an inlier, while $1$ denotes an outlier. $\mathbf{t}=\{t_{ij}\} \in \mathbb{N}$ and $\mathbf{p}=\{p_{ij}\} \in \{1, \ldots, P\}$ are the optional non-negative integer edge timestamp and edge type, respectively, where $P$ is the number of edge types.

\subsection{Problem Formulation}

In this paper, we focus on the task of graph outlier detection, which is formally defined as:

\begin{definition}[Graph Outlier Detection]
In this paper, we focus on node level outlier detection. Given a graph $\mathcal{G}$, graph outlier detection can be regarded as a binary classification task that learns a detector $D:v_i \rightarrow \{0, 1\}$ that classifies every node in $\mathcal{G}$ to an inlier (0) or an outlier (1).
\end{definition}

In the task of graph outlier detection, we aim to mitigate class imbalance by data augmentation:

\begin{definition}[Data Augmentation for Class Imbalance]
In outlier detection, the number of inliers is far more than the number of outliers, i.e., $|\{v_i \mid  y_i=0\}| \gg |\{v_i \mid  y_i=1\}|$. We aim to learn a parameterized data augmentation model to generate realistic and diverse synthetic graph $\hat{\mathcal{G}}=(\hat{\mathcal{V}}, \hat{\mathcal{E}}, \hat{\mathbf{X}}, \hat{\mathbf{y}}, \hat{\mathbf{t}}, \hat{\mathbf{p}})$, where $\hat{y_i}=1, \forall v_i \in \hat{\mathcal{V}}$. The synthetic graph $\hat{\mathcal{G}}$ is integrated with the original graph $\mathcal{G}$ to alleviate the class imbalance in the training of the graph outlier detector $D$.
\end{definition}

\section{Methodology}
\label{sec:method}

\begin{figure*}[t]
\centerline{
    \scalebox{1}{\includegraphics[width=\textwidth]{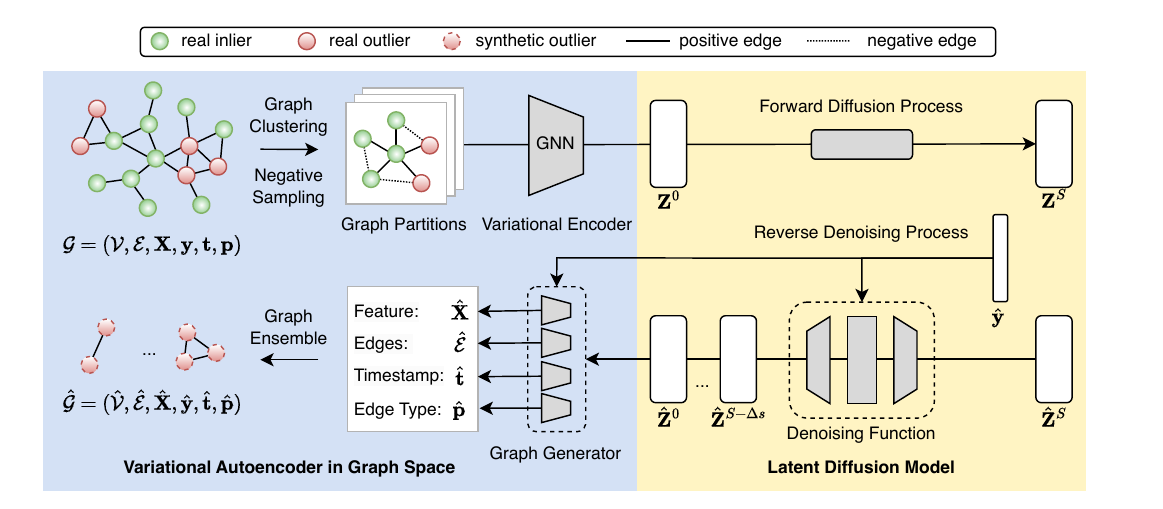}}}
    \caption{The architecture of proposed data augmentation method \method.}
\label{fig:godm}
\end{figure*}

In this section, we elaborate on the proposed data augmentation method \method. 

\subsection{Overview}

Figure~\ref{fig:godm} shows the architecture of \method. It starts by partitioning the input graph $\mathcal{G}$ with graph clustering algorithms and conducting negative sampling on edges to reduce memory consumption. Subsequently, each partitioned subgraph is encoded into a latent space representation denoted by $\mathbf{Z}^0$ through Variational Encoder based on graph neural networks (GNNs), followed by a forward diffusion $\mathbf{Z}^0 \rightarrow \mathbf{Z}^S$ in the latent space, where $S$ is the maximum step in the diffusion model. The reverse denoising process $\hat{\mathbf{Z}}^S \rightarrow \hat{\mathbf{Z}}^0$ iteratively estimates the noise and generates the latent embedding $\hat{\mathbf{Z}}^0$ from a predefined prior distribution $p(\hat{\mathbf{Z}}^S)$ using a denoising function, which is a multi-layer perceptron (MLP), conditioned on class labels $\hat{\mathbf{y}}$. Graph Generator employs the latent representations $\hat{\mathbf{Z}}^0$ to conditionally reconstruct the node features and graph structures alongside other information available in the original real graph. By ensembling generated information, we are able to obtain $\hat{\mathcal{G}}$ that contains the nodes that are statistically similar to the real outliers. This synthetic graph $\hat{\mathcal{G}}$, in concert with the original graph $\mathcal{G}$, is leveraged to the training of the downstream graph outlier detector $D$.

\subsection{Variational Encoder}
\label{enc}

Graph data is inherently complex and heterogeneous, containing both node feature $\mathbf{X}$ and graph structure $\mathcal{E}$, sometimes also partial node labels $\mathbf{y}$, temporal information $\mathbf{t}$, and edge types $\mathbf{p}$. However, current diffusion models are mainly designed for monotypic information (e.g., the magnitude of RGB channels in images). To bridge this gap, we adopt a GNN-based Variational Encoder $E: \mathcal{G} \rightarrow \mathbf{Z}$ to map different types of information in the graph space into a unified latent space.

\textbf{Node Feature}. To encode the node feature into the latent space, we take the feature of each node as the initial embedding for the input of the encoder, i.e., $\mathbf{h}_i^0 = \mathbf{x}_i$.

\textbf{Class Label}. As we only generate outliers by conditional generation, we encode node labels $\mathbf{y}=\{y_i\}$ into the initial node embedding by: $\mathbf{h}_i^0 = \mathbf{x}_i + \mathbf{w}^C_E \cdot y_i$, where $\mathbf{w}^C_E \in \mathbb{R}^{d}$ is a linear transformation.

\textbf{Graph Structure}. Graph neural networks (GNNs) have emerged as a profound architecture for learning graph data. GNNs efficaciously learn the node representations by encoding the graph topology and node feature simultaneously with message passing. In this paper, we take GraphSAGE~\cite{hamilton2017inductive} as an example. For each layer of GraphSAGE, each node updates its embedding by aggregating the message within the neighborhood:
\begin{equation}
\label{graphsage}
\mathbf{h}_i^l = \text{ACT}(\mathbf{W}^l \cdot \text{CAT}(\mathbf{h}_i^{l-1}, \text{AGG}(\{\mathbf{m}_{ij}^{l}, \forall e_{ij} \in \mathcal{E}\}),
\end{equation}

\noindent where $\mathbf{h}_i^l$ is the node embedding for node $v_i$ in the $l-$th layer, $\mathbf{m}_{ij}^{l}$ is the edge-wise message of edge $e_{ij}$, and ACT is a non-linear activation function. $\mathbf{W}^l \in \mathbb{R}^{d^l \times 2d^{l-1}}$ is the linear transformation, where $d^l$ and $d^{l-1}$ are the hidden dimension of $l$-th layer and $(l-1)$-th layer, respectively. While CAT represents concatenation, AGG denotes the aggregation function (e.g., mean). In vanilla GraphSAGE, an edge-wise message is the embedding of source node $v_j$ from the last layer $\mathbf{m}_{ij}^l = \mathbf{h}_j^{l-1}$.

\textbf{Edge Type}. However, sometimes, we have different types of edges in graphs, known as heterogeneous graphs or heterogeneous information networks \cite{shi2016survey, zhao2020network}. As different edge types can encapsulate various semantics, encoding edge types can be important for the downstream task. Therefore, we add the edge type information to the edge-wise message $\mathbf{m}_{ij}^l = \mathbf{h}_j^{l-1} + \mathbf{W}^P \cdot \phi(p_{ij})$, where $\mathbf{W}^P \in \mathbb{R}^{d^{l-1} \times P}$ is a linear transformation for edge type, and $\phi$ is the one-hot encoding.

\textbf{Edge Time}. In addition, temporal information is critical in time series applications \cite{deng2022graph}. When edge time is available, we are also able to encode the timestamp to the edge-wise message with temporal embedding: $\mathbf{m}_{ij}^l = \mathbf{h}_j^{l-1} + \mathbf{W}^T \cdot TE{(t_{ij}, \cdot)}$, where $\mathbf{W}^T \in \mathbb{R}^{d^{l-1} \times d^{l-1}}$ is the linear transformation for edge time, and $TE$ is the trigonometric temporal embedding~\cite{hu2020heterogeneous}.

By stacking multiple layers, GNNs are capable of encoding both node features and neighborhood graph structures into independent and identically distributed (i.i.d.) node embedding $\mathbf{Z}$, which can be further leveraged by Latent Diffusion Model. To ease the generation for Latent Diffusion Model, we use Variational Encoder, which outputs two different matrixes: 
\begin{equation}
\begin{aligned}
\label{eq:musigma}
\bm{\mu} = \text{GNN}_{\bm{\mu}}(\text{GNN}_{\text{shared}}(\mathcal{G})), \quad
\text{log}\bm{\sigma} = \text{GNN}_{\bm{\sigma}}(\text{GNN}_{\text{shared}}(\mathcal{G})),
\end{aligned}
\end{equation}

\noindent where $\bm{\mu}$ is the matrix of mean and log $\bm{\sigma}$ is the matrix of log standard deviation. While GNN$_{\text{shared}}$ is the shared GNN head, GNN$_{\bm{\mu}}$ and GNN$_{\bm{\sigma}}$ are specifically for $\bm{\mu}$ and $\bm{\sigma}$, respectively. Then, the latent space embedding $\mathbf{Z}$ can be obtained via the parameterization trick:
\begin{equation}
\label{eq:vae}
\mathbf{Z} = \bm{\mu} + \bm{\sigma} \cdot \bm \varepsilon, \quad \bm \varepsilon \sim \mathcal{N}(\bm{0}, \bm{I}),
\end{equation}

\noindent where $\mathcal{N}(\bm{0}, \bm{I})$ refers to a multivariate normal (or Gaussian) distribution with a mean of $\bm{0}$ and a covariance of $\bm{I}$, where $\bm{0}$ denotes a zero vector, and $\bm{I}$ represents a identity matrix. 

\subsection{Graph Generator}
\label{dec}

Graph Generator $G: \hat{\mathbf{Z}} \rightarrow \hat{\mathcal{G}}$ take the opposite process of Variational Encoder, generating $\hat{\mathcal{G}}=(\hat{\mathcal{V}}, \hat{\mathcal{E}}, \hat{\mathbf{X}}, \hat{\mathbf{y}}, \hat{\mathbf{t}}, \hat{\mathbf{p}})$ given the latent space embedding $\hat{\mathbf{Z}}$. Each row $\hat{\mathbf{z}}_i$ in the embedding $\hat{\mathbf{Z}}$ is corresponding to a generated node $\hat{v}_i$ in $\hat{\mathcal{V}}$.

\textbf{Class Label}. Recall that our goal is exclusively generating outliers with positive labels. Consequently, rather than generating class labels, we take the desired labels as an input for conditional graph generation. Specifically, the class label is added to the embedding: $\hat{\mathbf{z}}_i^C = \hat{\mathbf{z}}_i + \mathbf{w}^C_G \cdot \hat{y}_i$, where $\hat{\mathbf{z}}_i$ is the $i$-th row of $\hat{\mathbf{Z}}$ and $\hat{\mathbf{z}}_i^C$ is embedding with class condition. $\mathbf{w}^C_G \in \mathbb{R}^{d^L}$ is the linear transformation for the class label, and $d^L$ is the output dimension of the $L$-th layer of the encoder, i.e., the latent embedding dimension of $\hat{\mathbf{z}}_i$.

\textbf{Node Feature}. To generate the node feature $\hat{\mathbf{X}}$ from the latent space embedding, for the $i$-th row of $\hat{\mathbf{X}}$, we take the embedding with class condition $\hat{\mathbf{z}}_i^C$ as input:
\begin{equation}
\label{featuregen}
\hat{\mathbf{x}}_i = \mathbf{W}^F_G \cdot \hat{\mathbf{z}}_i^C,
\end{equation}

\noindent where $\mathbf{W}^F_G \in \mathbb{R}^{d \times d^L}$ is a learnable linear transformation matrix for node feature generation.

\textbf{Graph Structure}. The generation of graph structures denoted by the set of inferred edges $\hat{\mathcal{E}}$ requires link prediction between all pairwise combinations of nodes within the graph. The predicted edge score $\hat{e}_{ij}$ can be formulated as: $\hat{e}_{ij} =  \text{sigmoid}(\mathbf{w}^{E}_G \cdot \text{CAT}(\hat{\mathbf{z}}_i^C, \hat{\mathbf{z}}_j^C)), \forall (\hat{v}_i, \hat{v}_j) \in \mathcal{\hat{V}} \times \mathcal{\hat{V}}$, where $\mathbf{w}^E_G \in \mathbb{R}^{2d^L}$ is the linearn transformation for the edge generation. The generated edges are determined as follows:

\begin{equation}
\label{egen}
\hat{\mathcal{E}} = \{ (\hat{v}_i, \hat{v}_j) \mid  \hat{e}_{ij} \geq 0.5\} 
\end{equation}

\textbf{Edge Type}. With the generated edges in hand, we can predict the type of every generated edge:
\begin{equation}
\label{etypegen}
\hat{p}_{ij} =  \text{softmax}(\mathbf{W}^{P}_G \cdot \text{CAT}(\hat{\mathbf{z}}_i^C, \hat{\mathbf{z}}_j^C)), \forall (\hat{v}_i, \hat{v}_j) \in \mathcal{\hat{E}},
\end{equation}

\noindent where $\mathbf{W}^P_G \in \mathbb{R}^{P \times 2d^L}$ is the learnable linear transformation weight for the edge type prediction. 

\textbf{Edge Time}. Similarly, the timestamp of generated edges can be predicted as follows:
\begin{equation}
\label{etimegen}
\hat{t}_{ij} =  \mathbf{w}^{T}_G \cdot \text{CAT}(\hat{\mathbf{z}}_i^C, \hat{\mathbf{z}}_j^C), \forall (\hat{v}_i, \hat{v}_j) \in \mathcal{\hat{E}},
\end{equation}

\noindent where $\mathbf{w}^T_G \in \mathbb{R}^{2d^L}$ is the weight for the edge timestamp regression. 

\subsection{Latent Diffusion Model}
\label{ldm}

Utilizing Variational Encoder and Graph Generator, synthetic graph generation is already feasible. Nonetheless, the inherent complexity and heterogeneity of graph data pose a significant challenge when attempting a one-step estimation from a simple prior distribution, such as a Gaussian distribution, to the intricate target distribution. This challenge often leads to compromised generation quality and results in a marginal effect on the downstream graph outlier detection. Therefore, we integrate Latent Diffusion Model \cite{rombach2022high} to break down the estimation into a sequence of incremental steps. In each step, Latent Diffusion Model incrementally refines the distribution estimated, bridging the divergence between the simple prior and the intricate target distribution.

Latent Diffusion Model consists of a pair of processes. A fixed forward diffusion process $\{\mathbf{Z}(s)\}_{s=0}^S$, where $s \in [0, S]$ is a continuous diffusion step, which perturbs the original data $\mathbf{Z}(0)=\mathbf{Z}$ by the gradually adding Gaussian noise to obtain $\mathbf{Z}(S) \sim \mathcal{N}(\bm 0,\bm I)$, and a reverse denoising process $\{\hat{\mathbf{Z}}(s)\}_{s=0}^S$ employs a learned denoising function $\bm{\epsilon}_{\theta}$ to iteratively denoise the sampled noise from a simple prior distribution $\hat{\mathbf{Z}}(S)$ to obtain $\hat{\mathbf{Z}} = \hat{\mathbf{Z}}(0)$. In order to speed up the generation process, we adopt EDM \cite{karras2022elucidating} as our diffusion model. For the detailed design of forward process and reverse process, please refer to Appendix \ref{appx:dmf} and \ref{appx:dmr}, respectively.

\subsection{Training}
\label{train}

Conventional graph generative models are highly constrained in scalability. Typical graph generative models are working molecular graphs that are at hundreds node scale~\cite{de2018molgan}. However, outlier detection graphs are usually much larger, scaling to millions, even billions. Efficient training of generative models on large graphs requires special designs. We propose to apply negative sampling and graph clustering to improve the scalability of \method.

\textbf{Negative Sampling}. In the graph structure generation, if the training of the generator includes every pair of nodes, this leads to a computational complexity of $O(n^2)$. This parabolic complexity is catastrophic for large graphs (e.g., millions scale). Furthermore, it will result in a high imbalance in the training of the edge predictor itself. Negative sampling emerges as a crucial technique to reduce computational cost and alleviate the imbalance. To form a concise training set $\Bar{\mathcal{E}}$ for edge generator, apart from adding the positive edge, i.e., $\{\Bar{e}_{ij}=1 \mid e_{ij} \in \mathcal{E}\}$, we randomly select a negative edge (a pair of nodes that is not connected) corresponding to every positive edge, $\{\Bar{e}_{\Tilde{i}\Tilde{j}}=0 \mid e_{\Tilde{i}\Tilde{j}} \notin \mathcal{E} \}$. By this means, we reduce the complexity from $O(n^2)$ to $O(|\mathcal{E}|)$.

\textbf{Graph Clustering}. When the graph scales to millions of nodes, full-batch training becomes impractical, even with negative sampling. In addition, traditional neighbor sampling methods are well-suited in our case, as we need to reconstruct both the node feature and graph structure (i.e., edges). Specifically, node sampling cannot gather complete information for edge prediction, while edge sampling favors the nodes with a high degree. To address these challenges, we resort to graph clustering inspired by \cite{chiang2019cluster}. We first apply graph clustering algorithms (e.g., Metis \cite{karypis1997metis}) to divide the large graph into small partitions. Then, we form mini-batches with partitioned subgraphs and train \method on each mini-batch instead of the whole large graph.

Leveraging these two techniques, we are able to efficiently train \method in two steps. We first train Variational Encoder and Graph Generator to bridge the graph space and latent space. Then, we train the diffusion model in the latent space.

\textbf{Variational Encoder and Graph Generator}. By integrating Variational Encoder and Graph Generator, we are able to train both of them in a variational autoencoder (VAE) fashion. We reconstruct the node feature with MSE (mean squared error):
$\ell_{\mathbf{X}} = \frac{1}{|\mathcal{V}|}\sum_{v_i \in \mathcal{V}} \| \mathbf{x}_i - \hat{\mathbf{x}}_i \|_2^2$, where $\|\cdot\|_2$ denotes the L2 norm. Then, we predict the edges in the training edge set $\Bar{\mathcal{E}}$ with binary cross entropy loss: $\ell_{\mathcal{E}} =-\frac{1}{|\Bar{\mathcal{E}}|}\sum_{\Bar{e}_{ij} \in \Bar{\mathcal{E}}}(\Bar{e}_{ij}\log(\hat{e}_{ij})+(1-\Bar{e}_{ij}\log(1-\hat{e}_{ij})))$. Similarly, we reconstruct the timestamp with MSE loss: $\ell_{\mathbf{t}} = \frac{1}{|\mathcal{E}|}\sum_{e_{ij} \in \mathcal{E}} (t_{ij} - \hat{t}_{ij})^2$, and edge type with cross-entropy loss: $\ell_{\mathbf{p}} = -\frac{1}{|\mathcal{E}|}\sum_{e_{ij} \in \mathcal{E}} p_{ij} \log(\hat{p}_{ij})$.
We can obtain the total loss by the sum up of all reconstruction loss and the KL divergence loss:
\begin{equation}
\label{eq:vae_loss}
     \mathcal{L}_{\text{VAE}} = \omega_{\mathbf{X}} \ell_{\mathbf{X}} + \omega_{\mathcal{E}} \ell_{\mathcal{E}} + \omega_{\mathbf{t}} \ell_{\mathbf{t}} + \omega_{\mathbf{p}} \ell_{\mathbf{p}} + \beta \ell_{\rm kl},
\end{equation}

\noindent where $\omega_{\mathbf{X}}$, $\omega_{\mathcal{E}}$, $\omega_{\mathbf{t}}$, and $\omega_{\mathbf{p}}$ are the weights adjusting the values of each term in the reconstruction loss to a comparable scale. The $\ell_{\rm kl}$ is the KL divergence loss between the latent space embedding and prior distribution (e.g., Gaussian distribution). However, as we have an additional Latent Diffusion Model, we loosen this regularization on the latent embedding. Consequently, we use a small hyperparameter weight $\beta < 1$ to encourage the model to minimize reconstruction error while ensuring that the resultant embedding shape remains within the desired shape. The training of the VAE, consisting of Variational Encoder and Graph Generator, is presented in Algorithm \ref{alg:vae} in Appendix \ref{appx:alg}.

\textbf{Latent Diffusion Model}. Once obtain the well-trained Variational Encoder and Graph Generator, we start to train Latent Diffusion Model via denoising score matching.
We train a multi-layer perceptron (MLP) as the denoising function $\bm{\epsilon}_{\theta}$ for denoising score matching:
\begin{equation}
\min \mathbb{E}_{\mathbf{Z}^0 \sim p(\mathbf{Z}^0)} \mathbb{E}_{\mathbf{Z}^s \sim p(\mathbf{Z}^s|\mathbf{Z}^0)}  \Vert \bm{\epsilon}_{\theta}({\mathbf{Z}}^s_C, s) - {\bm{\varepsilon}} \Vert_2^2,
\end{equation}

\noindent where ${\mathbf{Z}}^s_C = \mathbf{Z}^s + \mathbf{y} \mathbf{W}^C, \mathbf{W}^C \in \mathbb{R}^{1\times d^L}$, encoding class label for conditional generation, and $\bm{\varepsilon}$ is the noise. More details about the training of Latent Diffusion Model are available in Appendix~\ref{appx:dmt}. The entire training process of Latent Diffusion Model is summarized in Algorithm \ref{alg:diffusion} in Appendix \ref{appx:alg}.

\subsection{Inference}
\label{sec:inf}

The well-trained \method is utilized in the inference process, augmenting real organic data by generated data. We first sample $\hat{\mathbf{Z}}^{{S}}$ from Gaussian distribution. Then, we iteratively denoise $\hat{\mathbf{Z}}^{{s_i}}$ to obtain $\hat{\mathbf{Z}}^{{s_{i-1}}}$ conditioned on $\hat{\mathbf{y}}=\bm 1$ via denoising function $\bm{\epsilon}_{\theta}$. The estimated $\hat{\mathbf{Z}}^{{0}}$ is fed into Graph Generator along with $\hat{\mathbf{y}}$ to generate synthetic graph $\hat{\mathcal{G}}$. Finally, the augmented graph $\mathcal{G}_{\text{sub}}$, which integrates the synthetic graph $\hat{\mathcal{G}}$ and the real organic graph $\mathcal{G}$, is used for training the downstream outlier detector $D$. Algorithm \ref{alg:inf} in Appendix \ref{appx:alg} presents the whole procedure of inference.

\subsection{Complexity Analysis}
\label{sec:ca}

To provide insights for handling large graphs which are common in real-world applications of graph outlier detection, we analyze the time efficiency and memory scalability of \method in Appendix \ref{appx:ca}.

\section{Experiments}
\label{sec:exp}

In this section, we systematically delineate the experiments to evaluate \method. 
The code implementation of \method is publicly available at: \url{https://github.com/kayzliu/godm}.

\input{tabs/tab1}

\subsection{Experimental Setups}
\label{sec:expsetup}

To benchmark \method and contemporary methods, we conduct experiments in a unified environment partially adapted from GADBench \cite{tang2023gadbench}. The implementation details are described in Appendix~\ref{appx:imp}.

\subsubsection{Datasets}

We use five different datasets from GADBench. Table~\ref{tab:data} in Appendix \ref{appx:data} provides the statistics of the datasets. Weibo, Elliptic, Tolokers, and Questions are homogeneous static graphs. DGraph is a heterogeneous temporal graph with different types of edges. The graphs vary in scale, ranging from thousands to millions of nodes. Detailed descriptions of datasets are available in Appendix~\ref{appx:data}.

\subsubsection{Baselines}

In our experiments, we evaluate \method against different types of baseline methods. We have three types of baseline methods.
For \textbf{general graph neural networks}, we choose:
GCN \cite{kipf2016semi}, SGC \cite{wu2019simplifying}, GIN \cite{xu2018powerful}, GraphSAGE \cite{hamilton2017inductive}, GAT \cite{velivckovic2018graph}, and GT \cite{shi2021masked}. For heterogeneous graphs, despite the rich literature, limited heterogeneous GNNs can cope with temporal information. As one of the representatives, we chose HGT \cite{hu2020heterogeneous} for our baseline. In addition to general GNNs, we also include eight \textbf{graph outlier detectors}: GAS \cite{li2019spam}, DCI \cite{wang2021decoupling}, PCGNN \cite{liu2021pick}, BernNet \cite{he2021bernnet}, GATSep \cite{zimek2014ensembles}, AMNet \cite{chai2022can}, and GHRN \cite{gao2023addressing}. Because the \textbf{data augmentation} for graph outlier detection is a relatively new research topic, there are limited baselines. We compare with DAGAD~\cite{liu2022dagad}, which is for homogeneous graphs. In addition, we introduce GOVAE, a variational autoencoder-only variant of \method, that can work on both homogeneous datasets and heterogeneous datasets. We drop Latent Diffusion Model and form a variational autoencoder with Variational Encoder and Graph Generator. The inference process is a direct one-step estimation from prior Gaussian distribution. Detailed descriptions of each baseline are available in Appendix \ref{appx:baseline}. Along with \method, we compare these three data augmentation methods for graph outlier detection in Table \ref{tab:comp} in Appendix \ref{appx:baseline}.

\subsubsection{Metrics}

We follow the extensive literature in graph outlier detection \cite{liu2022bond, tang2023gadbench} to comprehensively evaluate the performance of graph outlier detectors with three metrics robust to class imbalance: 
Receiver Operating Characteristic-Area Under Curve (AUC), Average Precision (AP), and Recall@k (Rec), where the value of k is set to the number of actual outliers present in the dataset. Appendix \ref{appx:metric} provides more details about the metrics used in the experiments.

\subsubsection{Open-Source Package} 

To enhance accessibility, we encapsulate our data augmentation into user-friendly API and make our code a plug-and-play package. This package is built upon PyTorch and PyTorch Geometric (PyG) frameworks. It accepts a PyG data object as input and returns the augmented graph as output. By inheriting the \texttt{transforms.BaseTransform} in PyG, we enable users to apply \method just like any other transformations supported in PyG, e.g., \texttt{ToUndirected}. We are also able to customize hyperparameters during initialization. Code Demo~\ref{lst:label} in Appendix \ref{appx:demo} gives an example of API usage. 
For distribution, we release our package on Python Package Index: \url{https://pypi.org/project/godm/}. Users can easily install \method in a command line by executing \texttt{\$ pip install godm}.

\subsection{Performamce on Graph Outlier Detection}
\label{sec:perf}

\begin{table}[t]
\begin{minipage}[t]{0.6\textwidth}
\input{tabs/tab2}
\end{minipage}
\hfill
\begin{minipage}[t]{0.38\textwidth}
\input{tabs/dgraph}
\end{minipage}
\end{table}

We start by examining how effective \method is in the task of graph outlier detection on homogeneous graphs. We report graph outlier detection performance on four homogeneous datasets in terms of AUC, AP, and Rec of different algorithms in Table \ref{tab:baseline}. In the table, the highest score in each metric is marked in bold, while the second-highest score is underlined. 
For DAGAD, we report the optimal performance of its two implementations with GCN and GAT. Although DAGAD shows superior performance compared to the standalone GCN and GAT, it fails to surpass more sophisticated detectors. On the other hand, \method and GOVAE together demonstrate significant enhancements in graph outlier detection performance across all datasets and metrics. Notably, on the Weibo dataset, where the performance metric is already relatively high (AUC of 98.66 for SGC), \method further elevates the performance, achieving an AUC of 99.57. A particularly noteworthy improvement is observed on Elliptic, where GODM enhances the Rec by over 20\% (an increase of 9.14) compared to GAS, which is the second-best performing algorithm. Although GOVAE exhibits competitive performance, its limited generation ability results in generally weaker performance compared to \method, especially on Elliptic.

Further investigations were carried out to understand the extent of improvement \method could bring to various graph outlier detection algorithms, specifically on Tolokers. We integrate \method on the top of all graph outlier detection algorithms and report the performance changes in terms of AUC, AP, and Rec in Table \ref{tab:boost}. The absolute performance is reported out of brackets, and the relative change is reported in the brackets. We can observe that \method enhances graph outlier detection performance for most of the algorithms. \method increases the AP of GAT by 7.99. For DCI and GAS, the only two algorithms that minorly decrease the performance, this could be attributed to suboptimal hyperparameter settings for these specific algorithms or the dataset. We believe the performance can be improved by careful hyperparameter tuning.

Moreover, in Table \ref{tab:dgraph}, we benchmark the graph outlier detector performance on DGraph, a large-scale dataset with different types of edges and temporal information. We include HGT along with variants of \method. For the w/o Feature, we randomly sample the node feature from the Gaussian distribution to replace the generated node feature in the synthetic graph. For the w/o Edge, all generated edges are substituted with random edges. For w/o Time, we give uniformly sampled timestamps to replace the generated timestamps. For the w/o Type, we use a random type sampled from a uniform distribution for all generated edges. According to Table \ref{tab:dgraph}, \method demonstrates enhanced performance across all metrics when compared to HGT. Furthermore, masking any generated information leads to a notable decline in performance, which underscores the significance of each aspect of the generated data. While GOVAE achieves commendable results, its performance still falls short of \method's, highlighting the superior effectiveness of Latent Diffusion Model.

\subsection{Generation Quality}
\label{sec:quality}

\begin{figure}[t]
    \vspace{-0.22in}
    \centering
    \begin{minipage}[t]{0.65\linewidth}
        \centering
        \includegraphics[width=\linewidth]{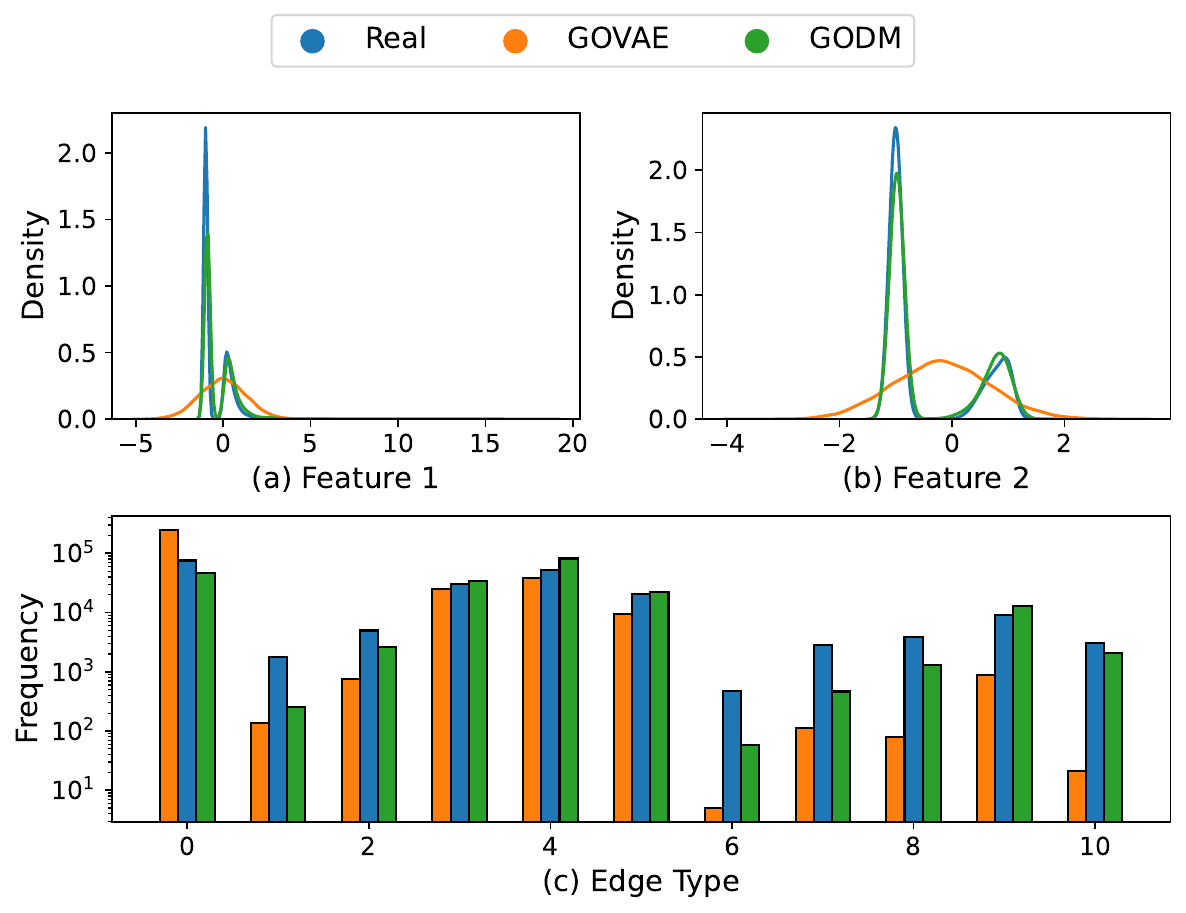}
        \caption{Visualization of single node feature density and edge type frequency of real data and synthetic data on DGraph.}
        \label{fig:den}
    \end{minipage}
    \hfill
    \begin{minipage}[t]{0.33\linewidth}
        \centering
        \includegraphics[width=\linewidth]{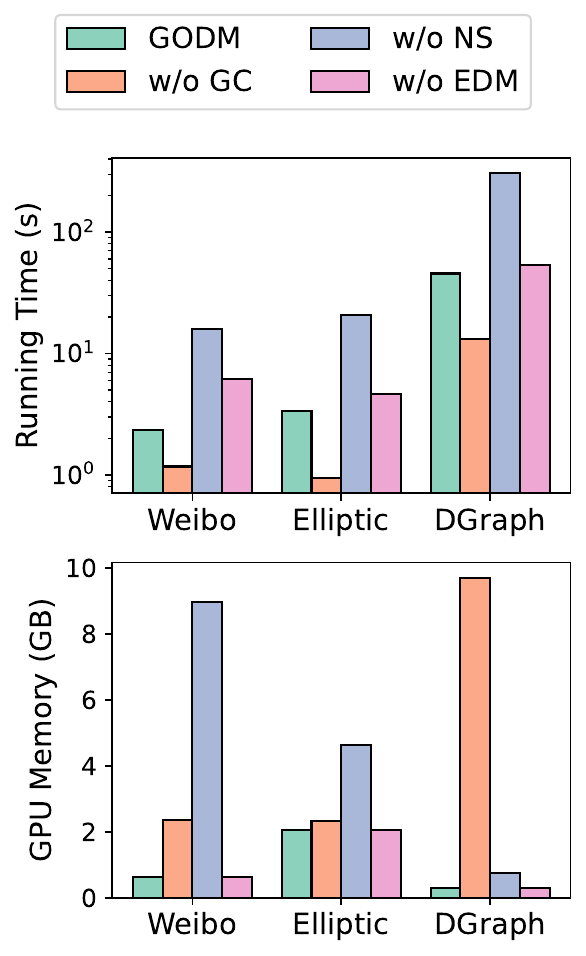}
        \caption{Running time and GPU memory consumption.}
        \label{fig:eff}
    \end{minipage}
    \vspace{-0.23in}
\end{figure}

To assess the quality of generated data, we conduct a case study on DGraph to compare real data and synthetic data from both GOVAE and \method, in Figure~\ref{fig:den}. Specifically, Figure \ref{fig:den} (a) and (b) illustrate the distribution density of a single dimension in node features. From the figures, we can see that the distributions of \method generated node features are close to the complex distributions of real data, while GOVAE can only generate Gaussian distributions, showcasing the generation ability of Latent Diffusion Model.
In addition, to evaluate generated edges, Figure~\ref{fig:den}~(c) shows the frequency across eleven different edge types. The edge type generated by \method has a more similar distribution to real data compared to GOVAE, particularly on the edge type 10. It further underscores the superiority of \method over GOVAE.

\subsection{Efficiency Study}
\label{sec:eff}

We further evaluate the efficiency of \method and its variants in terms of time and memory on graphs of various sizes, including Weibo, Elliptic, and DGraph. In order to demonstrate the efficiency of \method, we remove graph clustering (w/o GC), negative sampling (w/o NS), and EDM (w/o EDM) for evaluation. To simulate the real-world application, we measure the running time of unit epoch training for VAE and Latent Diffusion Model plus the inference process. For memory, we present the maximum active GPU memory usage within the whole data augmentation, as the GPU memory constraint is usually the bottleneck for machine learning systems. The results are shown in Figure \ref{fig:eff}.

From Figure \ref{fig:eff}, we can observe that employing a graph clustering algorithm significantly diminishes GPU memory utilization. Specifically, on DGraph, w/o GC results in a memory usage that is over 30$\times$ higher than \method. On the other hand, graph clustering turns full-batch training into mini-batch training, leading to an anticipated increase in running time. Additionally, negative sampling reduces both the running time and the memory usage, as the number of edges in the training set reduces from $O(n^2)$ to $O(|\mathcal{E}|)$. While EDM maintains consistent memory consumption, it saves running time by minimizing the number of sampling steps.

\section{Conclusion}
\label{sec:con}

In this paper, we introduce \method, a novel method for data augmentation in supervised graph outlier detection. GODM is designed to address the significant challenge of class imbalance, a pervasive issue in graph outlier detection.
\method leverages the power of latent diffusion models to synthesize high-fidelity graph data that is statistically similar to real outliers. 
Note that \method is model agnostic, which means it can be flexibly integrated with different downstream graph outlier detectors. We encapsulate \method into a plug-and-play package, making it accessible to the community. For future research, we list several promising directions in Appendix \ref{appx:future}.

\section*{Acknowledgement}

This work is supported in part by NSF under grants III-2106758, and POSE-2346158.

\bibliographystyle{unsrtnat}
\bibliography{reference}

\appendix

\section{Detailed Related Work}
\label{appx:rw}

This section summarizes the previous related works in three key areas, including graph outlier detection, class imbalance, and graph generative models.

\subsection{Graph Outlier Detection}

Graph outlier detection is a vital task in data mining and machine learning, aiming to identify anomalous structures (e.g., nodes, edges, subgraphs) in graphs. Due to its structural simplicity and broader applications, most research focuses on node-level graph outlier detection and can easily convert to edge level and graph level task~\cite{liu2024pygod}. 
Many studies focus on unsupervised settings, detecting outliers in the graph only based on the graph data itself without any ground truth labels \cite{ding2019deep, xu2022contrastive, liu2021anomaly}.
However, such unsupervised approaches may not align well with scenarios necessitating the detection of specific outlier types containing domain-specific knowledge. In this case, (semi-)supervised graph outlier detectors, which can learn ground truth labels, are better options. \cite{li2019spam} apply attention mechanism to detect spam review. \cite{dou2020enhancing, liu2020alleviating} alleviate the camouflage issue in fraud detection by message passing in selected neighborhoods. \cite{liu2021pick} further improves the neighbor selection in fraud detection. \cite{jin2021universal} builds the outlier detection graph via the KNN algorithm. \cite{wang2021decoupling} decouples the representation learning and anomaly detection, and \cite{zhu2020beyond} focuses on heterophily in graph outlier detection. Some other works detect outliers from a spectral perspective. \cite{he2021bernnet} learns arbitrary graph spectral filters via Bernstein approximation, and \cite{chai2022can} is capable of discerning both low-frequency and high-frequency signals, thereby adaptively integrating signals across a spectrum of frequencies. In \cite{tang2022rethinking}, the Beta kernel is employed to detect anomalies at higher frequencies using flexible and localized band-pass filters. \cite{gao2023addressing} addresses heterophily in graph outlier detection from a graph spectrum perspective. \cite{chang2024multitask, xu2024lego} alleviates the problem of limited supervision. \cite{wang2024uncertainty} summarizes uncertainty-based detection methods. \cite{chang2024enhancing} focuses on fair graph outlier detection. Despite the fruitful literature on graph outlier detection, most methods underrate a pivotal challenge of class imbalance. 

\subsection{Class Imbalance}

Conventional methods typically mitigate class imbalance by primitive techniques. GADBench uses reweighting \cite{tang2023gadbench}, assigning a higher weight to outliers in the loss function. Some works adopt upsampling to replicate outliers, which is mathematically equivalent to reweighting. \cite{dou2020enhancing} applies downsampling, reducing the number of normal instances and losing a large amount of valuable supervision information from normal instances. However, these tricks may not sufficiently address the underlying complexities associated with imbalanced datasets. Consequently, there is a compelling need for more sophisticated techniques that can effectively address this class imbalance to ensure the quality of graph outlier detection. \cite{liu2021pick} balances the neighborhood distribution by reducing edges between different classes and adding edges between the same classes but does not change the overall number of positive and negative nodes. \cite{liu2022dagad} adopts random perturbation for data augmentation to generate more samples. Other general graph machine learning studies apply interpolation-based data augmentation methods in latent space~\cite{zhao2021graphsmote}. However, these methods are not specifically designed for graph outlier detection, and interpolation is too naive to generate helpful instances. This paper seeks to build upon these foundational works, proposing a novel graph data augmentation algorithm via latent diffusion models that not only mitigate class imbalance effectively but are also scalable and efficient on large graphs, ensuring broader applicability in graph outlier detection.

\subsection{Graph Generative Models}

Recent advances in graph generative models have catalyzed a significant body of research, focusing primarily on the synthesis of realistic and structurally coherent graphs. \cite{kipf2016variational} proposes VGAE, leveraging variational autoencoder frameworks to learn latent representations of graph structures for generation purposes. \cite{you2018graphrnn} introduces GraphRNN, generating nodes and edges autoregressively with two recurrent neural networks. \cite{bojchevski2018netgan} applies a generative adversarial networks framework to generate graphs via random walks. \cite{luo2021graphdf} introduces the normalizing flow method for molecular graph generation. With the recent proliferation of works in image generation with diffusion models, some efforts have extended the generation power of diffusion models to graph-structured data. \cite{jo2022score} proposes a score-based generative modeling of graphs via the system of stochastic differential equations, while \cite{vignac2022digress} uses a discrete diffusion model to generate graphs. However, the potential of graph generative models to address the class imbalance in graph outlier detection remains largely underexplored. We propose to generate some synthetic outliers to mitigate this class imbalance.

\section{Detailed Design of Latent Diffusion Model}

We follow \cite{zhang2023mixed} for the detailed design of Latent Diffusion Model.

\subsection{Forward Diffusion Process}
\label{appx:dmf}
We construct a forward diffusion process $\{\mathbf{Z}(s)\}_{s=0}^S$, where $s \in [0, S]$ is a continuous diffusion step. 
In the diffusion process, $\mathbf{Z}(0)=\mathbf{Z}$ is the embedding from Variational Encoder, while $\mathbf{Z}(S) \sim \mathcal{N}(\bm 0,\bm I)$ is sampled from the prior distribution. According to \cite{song2020score}, the forward diffusion process can be written in stochastic differential equation (SDE) as:
\begin{equation}
\label{eq:forward}
{\rm d}\mathbf{Z} = \bm f(\mathbf{Z}, s){\rm d}s + g(s){\rm d} \bm{\omega}_t,
\end{equation}

\noindent where $\bm{\omega}_t$ is the standard Wiener process (i.e., Brownian motion). $\bm f(\cdot, s)$ and $g(\cdot)$ are the drift coefficient and the diffusion coefficient, respectively. The selection of $\bm f(\cdot, s)$ and $g(\cdot)$ vary between different diffusion models. $f(\cdot)$ is usually of the form $\bm f(\mathbf{Z}, s)  = f(s)\mathbf{Z}$. Thus, Equation \ref{eq:forward} can be written as: 
\begin{equation}
\label{eq:forward1}
{\rm d}\mathbf{Z} = f(s) \mathbf{Z} {\rm d}s + g(s){\rm d} \bm{\omega}_t.
\end{equation}

Let $\mathbf{Z}$ be a function of diffusion step $s$, i.e., $\mathbf{Z}^s=\mathbf{Z}(s)$. The diffusion kernel of Equation \ref{eq:forward1} can be represented in the conditional distribution of $\mathbf{Z}^s$ given $\mathbf{Z}^0$:
\begin{equation}\label{eq:kernel}
    p(\mathbf{Z}^s | \mathbf{Z}^0) = \mathcal{N}(a(s)\mathbf{Z}^0, a^2(s) \sigma^2(s) \bm I),
\end{equation}

\noindent where $a(s)$ and $\sigma(s)$ can be derived as: 
\begin{equation}
\label{eq:as_sigma}
a(s) = \exp{\int_0^sf(\xi)\rm d \xi}, \quad
\sigma(s)=\sqrt{\int_0^s g^2(\xi)/a^2(\xi)\rm d \xi}.
\end{equation}

\noindent Consequently, the formulation of the forward diffusion process is equivalent to the definition of diffusion kernels characterized by both $a(s)$ and $\sigma(s)$.

Denoising diffusion probabilistic models (DDPM)~\cite{ho2020denoising} can be seen as discretizations of the variance preserving SDE with $a(s)=\sqrt{1-\beta(s)}$ and $\sigma(s) = \sqrt{\beta(s)/{(1-\beta(s))}}$, as $a^2(s) + a^2(s)\sigma^2(s) = 1$. However, in order to achieve more efficient generation in \method, we adopt EDM \cite{karras2022elucidating}, which belongs to variance exploding SDE. Variance exploding SDE set $a(s)=1$, which implies that noise is directly added to the data instead of being blended through weighting. In this case, the variance of the noise (the noise level) is exclusively determined by $\sigma(s)$. In EDM, a linear noise level $\sigma(s) = s$ is applied. Therefore, the diffusion kernel can be written as:
\begin{equation}
    p(\mathbf{Z}^s | \mathbf{Z}^0) = \mathcal{N}(\bm{0}, \sigma^2(s)\bm I),
\end{equation}

\noindent and the forward diffusion process can be formulated as:
\begin{equation}
\mathbf{Z}^s = \mathbf{Z}^0 + \sigma(s) \bm{\varepsilon}, \ \bm{\varepsilon} \sim \mathcal{N}(\bm{0}, \bm{I}).
\end{equation}

\subsection{Reverse Denoising Process}
\label{appx:dmr}
As derived in \cite{song2020score}, the reverse denoising process is formulated as the reverse SDE of Equation \ref{eq:forward}:
\begin{equation}
\label{eq:backward}
{\rm d}\mathbf{Z} = [\bm f(\mathbf{Z},s) -g^2(s)\nabla_{\mathbf{Z}} \log p_s(\mathbf{Z})] {\rm d}s + g(s) {\rm d} \bm{\omega}_s,
\end{equation}

\noindent where $\nabla_{\mathbf{Z}} \log p_s(\mathbf{Z})$ is the score function of $\mathbf{Z}$. In EDM, given $a(s) = 1$ and Equation \ref{eq:as_sigma},
\begin{equation}
\bm f(\mathbf{Z}, s) = f(s) \mathbf{Z} = \bm{0}, \quad
g(s) = \sqrt{2\sigma(s) \dot{\sigma}(s)},
\end{equation}

\noindent where $\dot{\sigma}$ denotes the first order derivative of $\sigma$. With $\bm f(\mathbf{Z}, s)$ and $g(s)$, we are able to obtain:
\begin{equation}
\label{eq:backward1}
{\rm d}\mathbf{Z} = -2\dot{\sigma}(s) \sigma(s) \nabla_{\mathbf{Z}} \log p_s(\mathbf{Z}) {\rm d}s + \sqrt{2\dot{\sigma}(s) \sigma(s)} {\rm d} \bm{\omega}_s,
\end{equation}

\noindent where the noise level $\sigma(s)=s$.

\subsection{Training of Latent Diffusion Models}
\label{appx:dmt}
To solve the SDE in Equation \ref{eq:backward1}, we need to obtain the score function $\nabla_{\mathbf{Z}} \log p_s(\mathbf{Z})$, which is as intractable as $p_s(\mathbf{Z})=p(\mathbf{Z}^s)$. However, the conditional distribution $p(\mathbf{Z}^s|\mathbf{Z}^0)$ is tractable. From Equation \ref{eq:kernel}, we obtain its analytical solution:
\begin{equation}
\nabla_{\mathbf{Z}} \log p(\mathbf{Z}^s|\mathbf{Z}^0)=- \frac{\bm{\varepsilon}}{a(s)\sigma(s)}.
\end{equation}

\section{Algorithms}
\label{appx:alg}

In this section, we provide algorithms used in \method. Algorithm \ref{alg:vae} gives the training process of VAE. Algorithm \ref{alg:diffusion} shows the training process of Latent DIffusion Model. Algorithm \ref{alg:inf} presents the inference process.

\begin{algorithm}[H]
    \centering
    \caption{\method: Training of VAE}\label{alg:vae}
    \begin{algorithmic}[1]
        \REQUIRE Input graph $\mathcal{G} = ( \mathcal{V}, \mathcal{E}, \mathbf{X}, \mathbf{y}, \mathbf{t}, \mathbf{p})$
        \ENSURE Node representation $\mathbf{Z}$, well-trained Variational Encoder $E$ and Graph Generator $G$
        \STATE Partition graph $\mathcal{G}$ into a set of subgraphs $\{\mathcal{G}_\text{sub}\}$
        \FOR{each $\mathcal{G}_\text{sub}$}
        \STATE Get $\bm \mu$ and log$\bm \sigma$ via Equation \ref{eq:musigma}
        \STATE Reparameterization of $\mathbf{Z}$ via Equation \ref{eq:vae}
        \STATE Generate $\hat{\mathbf{X}}$, $\hat{\mathcal{E}}$, $\hat{\mathbf{t}}$, $\hat{\mathbf{p}}$ via Equation \ref{featuregen}, \ref{egen}, \ref{etimegen}, \ref{etypegen}
        \STATE Calculate loss via Equation \ref{eq:vae_loss}
        \STATE Update Variational Encoder and Graph Generator parameters via the Adam optimizer
        \ENDFOR
    \end{algorithmic}
\end{algorithm}

\begin{algorithm}[H]
    \centering
    \caption{\method: Training of Latent Diffusion Model}\label{alg:diffusion}
    \begin{algorithmic}[1]
        \REQUIRE Node representation $\mathbf{Z}$
        \ENSURE Well-trained denoising function $\bm{\epsilon}_{\theta}$
        \STATE Sample the embedding $\mathbf{Z}^0$ from $p(\mathbf{Z}) = p(\bm{\mu})$
        \STATE Sample diffusion steps $s$ from $p(s)$ then get $\sigma (s)$
        \STATE Sample noise vectors ${\bm{\varepsilon}} \sim \mathcal{N}(\bm{0}, \sigma^2(s) \bm{I})$
        \STATE Get perturbed data ${\mathbf{Z}}^{s} = \mathbf{Z}^0 + {\bm{\varepsilon}}$
        \STATE Calculate loss $\ell(\theta) =  \| \bm{\epsilon}_{\theta}({\mathbf{Z}}^s_C, s)-\bm{\varepsilon}\|_2^2$
        \STATE Update the network parameter $\theta$ via Adam optimizer
    \end{algorithmic}
\end{algorithm}

\begin{algorithm}[ht]
    \centering
    \caption{\method: Inference}\label{alg:inf}
    \begin{algorithmic}[1]
        \REQUIRE Graph generator $G$ and denoising function $\bm{\epsilon}_{\theta}$
        \ENSURE Augmented graph $\mathcal{G}_{\text{aug}}$
        \STATE Sample $\hat{\mathbf{Z}}^{{S}} \sim \mathcal{N}(\bm{0}, \sigma^2({S})\bm{I})$
        \FOR{$i = \max, \cdots, 1$}
        \STATE $\nabla_{\hat{\mathbf{Z}}^{s_i}}\log p(\hat{\mathbf{Z}}^{s_i}) = -\bm{\epsilon}_{\theta}(\hat{\mathbf{Z}}^{s_i}_C, {s_i}) / \sigma({s_i})$
        \STATE get $\hat{\mathbf{Z}}^{s_{i-1}}$ via solving the SDE in Equation \ref{eq:backward1}
        \ENDFOR
        \STATE Generate $\hat{\mathcal{G}} = G(\hat{\mathbf{Z}}^0, \hat{\mathbf{y}})$
        \STATE $\mathcal{G}_{\text{aug}} = \text{batch}(\mathcal{G}, \hat{\mathcal{G}})$
    \end{algorithmic}
\end{algorithm}

\section{Complexity Analysis}
\label{appx:ca}

In this section, we meticulously analyze and elucidate the time efficiency and memory scalability of \method in training and inference for both VAE (Variational Encoder and Graph Generator) and Latent Diffusion Model.

\subsection{Time Efficiency}
In the training of VAE, the major bottleneck comes from the graph structure generation in Graph Generator. This generation originally requires link prediction for every pair of nodes in the graphs, which is $O(n^2)$. By graph clustering, we divide the large graph into relatively small partitions for mini-batch training. We denote the average partition size (i.e., batch size) by $b$, so the number of partitions is $\frac{n}{b}$. Then, the complexity is reduced to $O(\frac{n}{b} b^2)=O(nb)$, as only node pairs within each partition are predicted. We apply the negative sampling to further reduce the complexity to $O(|\mathcal{E}|)$, as we only sample a negative edge for each positive edge. For Latent Diffusion Model, we only sample a fixed number of diffusion steps to train the denoising function. As a result, assuming the number of latent dimensions is a constant, the complexity is $O(n)$. Thus, the total training complexity is $O(n + |\mathcal{E}|)$.
For the inference, we first take $S$ steps sampling in Latent Diffusion Model, which is $O(nS)$, and predict edges between node pairs in each partition, which is $O(nb)$. Therefore, the total inference complexity is $O(nS+nb)$.

\subsection{Memory Scalability}
The vanilla VAE requires memory consumption of $O(n^2)$, which is infeasible for large graphs. To reduce memory consumption, we adopt mini-batch training on graph partitions and negative sampling for link prediction. The complexity of VAE training can be obtained by the number of edges divided by the number of partitions $O(|\mathcal{E}|/\frac{n}{b})=O(\frac{|\mathcal{E}|b}{n})$. Latent Diffusion Model only requires a memory of $O(b)$. In the inference process, the VAE performs link prediction for all node pairs in a batch, leading to a memory complexity of $O(b^2)$, and Latent Diffusion Model only demands $O(b)$ for each batch. Therefore, the total inference process needs $O(b^2)$ of memory.

\section{Implementation Details}
\label{appx:imp}

We modified GADBench \cite{tang2023gadbench} to benchmark graph outlier detection performance.

\textbf{Environment}. The key libraries and their versions used in experiments are as follows: Python 3.9, CUDA 11.8, PyTorch 2.0.1~\cite{paszke2019pytorch}, PyG 2.4.0 \cite{fey2019fast}, DGL 1.1.2~\cite{wang2019deep}, and PyGOD 1.0.0 \cite{liu2024pygod}.

\textbf{Hardware}. All of our experiments were performed on a Linux server with an AMD EPYC 7763 64-core CPU, 192GB RAM, and an NVIDIA RTX A40 GPU with 48GB memory.

\textbf{Hyperparameters}. \method is implemented with default hyperparameters. For Variational Encoder, we use one layer GraphSAGE for $\text{GNN}_{\text{shared}}$, $\text{GNN}_{\bm{\mu}}$, and $\text{GNN}_{\bm{\sigma}}$, respectively. We set the hidden dimension to the largest power of 2 that is no greater than the feature dimension of the dataset divided by 2. To avoid extensive hyperparameter search, we set the weights of reconstruction loss $\omega_{\mathbf{X}}$, $\omega_{\mathcal{E}}$, $\omega_{\mathbf{t}}$, and $\omega_{\mathbf{p}}$ to roughly balance their absolute value only, which are 1, 0.5, 1, and 0.3, respectively. The weight for KL-divergence $\beta$ is 0.001. For the diffusion model, we use the dimension twice as the hidden dimension in VAE. We use five layers of MLP as the denoising function following the detailed design in \cite{zhang2023mixed}. For training, we adopt the Adam optimizer with a learning rate of 0.001 without weight decay to train the VAE and the diffusion model for 100 epochs, respectively, and apply early stopping with a patience of 50. The negative sampling ratio is 1, and the approximate graph partition size is 2048. For inference, we use 50 steps of diffusion and generate the same amount of synthetic outliers as the number of real organic outliers in the training set. For all the hyperparameters in graph outlier detection, we apply the default setting in GADBench. We report the graph outlier detection performance of \method with the optimal downstream graph outlier detector.

\section{Description of Datasets}
\label{appx:data}

\input{tabs/data}

This section describes the datasets used in the experiments. Table~\ref{tab:data} provides the statistics of the datasets.
In the table, \#Nodes stands for the number of nodes, and \#Edges stands for the number of edges. \#Feature denotes the raw feature dimension, i.e., the number of node attributes. The Outlier column represents the outlier ratio in the label, indicating the extent of class imbalance. The detailed descriptions for each dataset used in the experiments are as follows:

\textbf{Weibo} \cite{zhao2020error}: This dataset involves a graph delineating the associations between users and their corresponding hashtags from the Tecent-Weibo platform, consisting of 8,405 users and a collection of 61,964 hashtags. Activities within this dataset are regarded as suspicious (i.e., outliers) if they consist of a pair of posts occurring within narrowly defined temporal intervals (e.g., 60 seconds). Users engaging in at least five incidents of such behavior are classified under the suspicious category, in contrast to the remainder designated as benign. Following this classification criterion, the dataset contains 868 users identified as suspicious and 7,537 as benign. The primary feature vector includes geolocation data for each micro-blog entry and a representation utilizing the bag-of-words model.

\textbf{Tolokers} \cite{platonov2022critical}: The dataset is obtained from the Toloka crowdsourcing platform. It is composed of nodes corresponding to individual workers who have engaged in at least one out of thirteen specified projects. Edges are established between pairs of workers who have concurrently contributed to an identical task. The primary objective is to predict the likelihood of a worker having received a ban in any one of the projects. The features attributed to each node are constructed utilizing the worker's personal profile information in conjunction with their task-related metrics.

\textbf{Questions} \cite{platonov2022critical}: The dataset is derived from Yandex Q, a question-answering platform. The nodes represent users, and an edge is established between two nodes to denote the scenario wherein one user has responded to another’s inquiry within the temporal bounds of one year, stretching from September 2021 to August 2022. This dataset is specifically curated to encapsulate the engagement of users who exhibit interest in the medical topic. The target for this dataset is to predict the likelihood of users’ continued activity on the platform by the conclusion of the observed period. The average FastText embeddings of the lexical content present in the users’ descriptions are used for node features.

\textbf{Elliptic} \cite{weber2019anti}: The dataset comprises a graph of 203,769 nodes representing Bitcoin transactions, connected through 234,355 edges representing payment flows, along with 166 distinctive node attributes. It correlates Bitcoin transactional data with corresponding real-world entities that are categorized as lawful, including exchanges, wallet services, mining operations, and legitimate services, as well as unlawful categories, including scams, malicious software, terrorist-related organizations, ransomware operations, and fraudulent investment activities known as Ponzi schemes.

\textbf{DGraph} \cite{huang2022dgraph}: DGraph is a large-scale graph with different edge types and temporal information supplied by Finvolution Group. It includes around 3 million nodes, 4 million dynamic edges, and 1 million node labels. The nodes represent user accounts within a financial organization offering personal loan services, while an edge between two nodes indicates that one account has designated the other as an emergency contact. Nodes classified as fraud correspond to users displaying delinquent financial conduct. For those accounts with borrowing records, outliers are identified as accounts with a history of overdue payments, while inliers are those without such a history. Additionally, the dataset includes 17 node features derived from user profile information.

\section{Description of Baselines}
\label{appx:baseline}

This section provides detailed descriptions of the baselines in the experiments.

\subsection{General Graph Neural Network}

\textbf{GCN} (Graph Convolutional Networks) \cite{kipf2016semi}: GCN is the seminal work that applies convolution operation on graph data. It propagates the information of a node to its neighbors, thus enabling the network to develop a representation for each node that reflects its local neighborhood context.

\textbf{SGC} (Simplified Graph Convolution) \cite{wu2019simplifying}: This variant of GCN leverages Chebyshev polynomials to approximate the spectral graph convolution operator. This strategy allows the model to encompass both local and global graph structures, enhancing its scalability for handling larger graphs.

\textbf{GIN} (Graph Isomorphism Network) \cite{xu2018powerful}: GIN is a form of GNN that effectively captures graph structures while maintaining graph isomorphism. It achieves this by generating consistent embeddings for structurally identical graphs, which is permutation invariant.

\textbf{GraphSAGE} (Graph Sample and Aggregate) \cite{hamilton2017inductive}: GraphSAGE presents a general inductive learning approach where node embeddings are generated through the sampling and aggregation of features from a node's immediate neighborhood.

\textbf{GAT} (Graph Attention Networks) \cite{velivckovic2018graph}: GAT incorporates the attention mechanism within the GNN framework. It dynamically assigns varying importance to different nodes during the information aggregation process, focusing the model's learning on the most relevant parts of the neighborhood.

\textbf{GT} (Graph Transformer) \cite{shi2021masked}: Drawing inspiration from the Transformer model in neural networks, GT adapts these principles for graph-structured data. It utilizes masks in the self-attention mechanism to capitalize on the inherent structure of graphs, thus boosting the model's efficiency.

\textbf{HGT} (Heterogeneous Graph Transformer) \cite{hu2020heterogeneous}: HGT is designed to address the challenges of modeling heterogeneous graphs. It introduces node- and edge-type dependent parameters for heterogeneous attention mechanisms for each type of edge.

\subsection{Graph Outlier Detector}

\textbf{GAS} (GCN-based Anti-Spam) \cite{li2019spam}: GAS is an attention-based spam review detector, extending the capabilities of GCN to process heterogeneous and heterophilic graphs. It employs the KNN algorithm to align with the structure of each graph.

\textbf{DCI} (Deep Cluster Infomax) \cite{wang2021decoupling}: DCI is a self-supervised learning strategy that separates the learning of node representations from outlier detection. It addresses discrepancies between node behavioral patterns and their label semantics by clustering, thus capturing intrinsic graph properties in focused feature spaces.

\textbf{PCGNN} (Pick and Choose Graph Neural Network) \cite{liu2021pick}: Tailored for imbalanced GNN learning in fraud detection scenarios, PCGNN uses a label-balanced sampler for node and edge selection during training. This results in a more balanced label distribution within the induced subgraph.

\textbf{BernNet} \cite{he2021bernnet}: BernNet is a GNN variant offering a robust approach to designing and learning arbitrary graph spectral filters. It utilizes an order-K Bernstein polynomial approximation for estimating filters over the normalized Laplacian spectrum, catering to a variety of graph structures.

\textbf{GATSep} \cite{zimek2014ensembles}: Designed to optimize learning on heterophily graphs, GAT-sep merges key design elements like ego- and neighbor-embedding separation, higher-order neighborhood processing, and combinations of intermediate representations.

\textbf{AMNet} (Adaptive Multi-frequency GNN) \cite{chai2022can}: AMNet is structured to capture signals across both low and high frequencies by stacking multiple BernNets and adaptively integrating signals from different frequency bands.

\textbf{BWGNN} (Beta Wavelet Graph Neural Network) \cite{tang2022rethinking}: BWGNN addresses the "right-shift" phenomenon on outliers. It uses the Beta kernel to address higher frequency anomalies through spatially/spectrally localized band-pass filters.

\textbf{GHRN} (Graph Heterophily Reduction Network) \cite{gao2023addressing}: GHRN tackles the issue of heterophily in the spectral domain for graph outlier detection. This approach focuses on pruning inter-class edges to enhance the representation of high-frequency components in the graph spectrum.

\subsection{Data Augmentation}

\textbf{DAGAD} (Data Augmentation for Graph Anomaly Detection)~\cite{liu2022dagad}: DAGAD incorporates three modules to augment the graph, including an information fusion module for representation learning, a data augmentation module to enrich the training set with synthetic samples, and an imbalance-tailored learning module to distinguish between minority anomalous class and majority normal class.

\textbf{GOVAE}: In this variant, we drop Latent Diffusion Model and form a variational autoencoder with Variational Encoder and Graph Generator. The inference process is a direct one-step estimation from prior Gaussian distribution.

Along with \textbf{\method}, we compare three data augmentation methods in Table \ref{tab:comp}, where the flexibility represents whether the data augmentation is model agnostic. Heterogeneous and time indicate whether the method supports these two types of information. The diffusion column indicates whether the method adopts diffusion models.

\input{tabs/comp}

\section{Description of Metrics}
\label{appx:metric}

\textbf{AUC} (Receiver Operating Characteristic-Area Under Curve): AUC quantifies the area beneath the Receiver Operating Characteristic Curve, which is constructed by plotting the true positive rate against the false positive rate across varied determined threshold levels. An AUC of 1 indicates flawless predictive accuracy, whereas an AUC of 0.5 indicates an absence of discriminative power, equivalent to random guessing. This metric is favored over accuracy for evaluating outlier detection models due to its robustness against class imbalance prevalent within the class distributions.

\textbf{AP} (Average Precision): AP offers a comprehensive summary of the precision-recall curve, represented as the weighted average of precision values attained at each threshold, utilizing the increment in recall from the preceding threshold as the weight. This metric offers a balance between recall and precision, with a higher AP signifying a lower rate of both false-positive rate (FPR) and false-negative rate (FNR). For outlier detection applications, e.g., fraud detection, both FPR and FNR are critical, as misclassification could either cause financial loss or harm normal user experience.

\textbf{Rec} (Recall@k): Considering the minority of outliers relative to the abundance of normal instances in datasets, Recall@k is proposed as a measure of how well the detectors rank outliers over the normal samples. The value of k is set to the number of actual outliers present in the dataset. Recall@k is then determined by the ratio of the true outliers within the top k-ranked samples to k. A maximal Recall@k value of 1 means a model perfectly ranks all outliers over normal samples, while minimal value of 0 means none of the outliers is ranked at top k.

\section{API Demo}
\label{appx:demo}

\input{tabs/api}

\section{Future Directions}
\label{appx:future}

For future research, here are several promising directions we may explore to further enhance \method and expand its applicability in the domain of graph outlier detection: 
\begin{itemize}
    \item \textbf{Diffusion in graph sapce}. Considering the heterogeneity and scalability of the graphs for outlier detection applications, \method currently employs a diffusion model in latent space. \cite{li2023graphmaker} starts to generate large graphs directly in graph space. Future research could explore the application of diffusion models on heterogeneous information directly in graph space, potentially enhancing the model's ability to capture complex graph structures.
    \item \textbf{Collaboration with outlier detection}. Currently, \method is model agnostic, meaning the training of GODM and downstream outlier detection tasks are asynchronous. On the other hand, integrating the supervision signals from both the generation task and the downstream outlier detection in a unified training framework could be mutually beneficial. Self-training offers a promising paradigm for this integration of the supervision signals~\cite{liu2022confidence}.
    \item \textbf{Enhancing generative models}. Although the current diffusion model in \method has shown impressive capabilities, there are opportunities for further improvement. This could be achieved by employing more expressive denoising functions, such as transformers \cite{peebles2023scalable}, or by exploring alternative generative models like energy-based models \cite{zhao2016energy} and or normalizing flows \cite{rezende2015variational}, which might offer different advantages over diffusion models.
\end{itemize}

\end{document}

%% file: dfn.tex
\definecolor{dkred}{rgb}{0.5,0,0}
\definecolor{dkgreen}{rgb}{0,0.6,0}
\definecolor{gray}{rgb}{0.5,0.5,0.5}
\definecolor{mauve}{rgb}{0.58,0,0.82}

%% file: tabs/tab1.tex
\begin{table*}[ht]
\centering
\caption{Performance in AUC, AP, and Rec (\%) on four datasets.}
\label{tab:baseline}
\scalebox{0.75}{
\begin{tabular}{@{}lccclccclccclccc@{}}
\toprule
\textbf{Dataset}   & \multicolumn{3}{c}{\textbf{Weibo}}               & \textbf{} & \multicolumn{3}{c}{\textbf{Tolokers}}            &  & \multicolumn{3}{c}{\textbf{Questions}}           &  & \multicolumn{3}{c}{\textbf{Elliptic}}            \\ \cmidrule(lr){2-4} \cmidrule(lr){6-8} \cmidrule(lr){10-12} \cmidrule(l){14-16} 
\textbf{Metric}    & AUC            & AP             & Rec            &           & AUC            & AP             & Rec            &  & AUC            & AP             & Rec            &  & AUC            & AP             & Rec            \\ \midrule
\textbf{GCN \cite{kipf2016semi}}       & 98.11          & 93.48          & 89.34          &           & 74.69          & 42.88          & 42.06          &  & 69.81          & 12.54          & 16.99          &  & 82.68          & 22.23          & 27.61          \\
\textbf{SGC \cite{wu2019simplifying}}       & 98.66          & 92.46          & 87.90          &           & 70.67          & 38.03          & 35.98          &  & 69.88          & 10.13          & 15.62          &  & 73.02          & 11.44          & 9.14           \\
\textbf{GIN \cite{xu2018powerful}}       & 97.47          & 92.67          & 87.90          &           & 74.05          & 36.57          & 36.76          &  & 67.76          & 12.30          & 18.36          &  & 84.38          & 29.66          & 35.64          \\
\textbf{GraphSAGE \cite{hamilton2017inductive}} & 96.54          & 89.25          & 86.17          &           & 79.42          & 48.65          & 46.42          &  & 71.69          & 17.63          & 21.10          &  & 85.31          & 37.52          & 36.20          \\
\textbf{GAT \cite{velivckovic2018graph}}       & 94.08          & 90.25          & 86.74          &           & 77.26          & 43.14          & 43.30          &  & 70.33          & 14.51          & 17.26          &  & 84.42          & 23.43          & 27.42          \\
\textbf{GT \cite{shi2021masked}}        & 97.06          & 91.44          & 87.03          &           & 79.24          & 46.22          & 46.57          &  & 70.83          & 16.14          & 20.27          &  & 87.14          & 29.91          & 38.97          \\
\hline
\textbf{GAS \cite{li2019spam}}       & 94.88          & 90.70          & 86.74          &           & 76.91          & 47.35          & 45.02          &  & 64.50          & 13.61          & 17.53          &  & {\ul 87.81}    & 40.03          & {\ul 44.78}    \\
\textbf{DCI \cite{wang2021decoupling}}       & 93.90          & 87.78          & 83.86          &           & 75.98          & 39.85          & 40.19          &  & 67.95          & 14.58          & 19.18          &  & 81.93          & 27.63          & 33.15          \\
\textbf{PCGNN \cite{liu2021pick}}     & 90.89          & 84.57          & 79.83          &           & 72.18          & 37.52          & 36.76          &  & 68.38          & 14.79          & 16.99          &  & 86.50          & {\ul 42.66}    & 43.77          \\
\textbf{GATSep \cite{zimek2014ensembles}}    & 96.72          & 91.55          & 89.05          &           & 79.63          & 46.08          & 46.73          &  & 69.96          & 15.98          & 19.18          &  & 83.89          & 21.46          & 21.15          \\
\textbf{BernNet \cite{he2021bernnet}}   & 93.85          & 88.00          & 85.30          &           & 76.20          & 42.20          & 42.21          &  & 70.80          & 16.04          & 17.53          &  & 82.01          & 20.52          & 23.55          \\
\textbf{AMNet \cite{chai2022can}}     & 95.88          & 89.74          & 85.59          &           & 75.83          & 42.66          & 41.90          &  & 69.71          & 17.02          & 19.18          &  & 80.06          & 16.73          & 17.17          \\
\textbf{BWGNN \cite{tang2022rethinking}}     & 98.29          & 92.72          & 84.73          &           & 80.15          & 49.65          & 47.35          &  & 69.47          & 16.24          & 18.63          &  & 84.32          & 22.56          & 26.50          \\
\textbf{GHRN \cite{gao2023addressing}}      & 97.21          & 92.67          & 88.18          &           & 79.80          & 49.50          & 48.29          &  & 68.24          & 16.24          & 18.63          &  & 85.36          & 24.01          & 30.29          \\
\hline
\textbf{DAGAD \cite{liu2022dagad}}     & 98.54          & 83.36          & 90.78          &           & 77.69          & 33.94          & 44.39          &  & 71.21          & 6.88           & 20.55          &  & 85.62          & 26.18          & 40.54          \\
\textbf{GOVAE (ours)}     & {\ul 99.46}    & {\ul 96.84}    & \textbf{93.08} &           & {\ul 83.42}    & \textbf{53.85} & {\ul 52.49}    &  & {\ul 75.73}    & {\ul 19.13}    & {\ul 23.84}    &  & 83.93          & 36.66          & 42.66          \\
\textbf{GODM (ours)}      & \textbf{99.57} & \textbf{97.54} & \textbf{93.08} &           & \textbf{83.46} & {\ul 52.95}    & \textbf{52.96} &  & \textbf{76.84} & \textbf{20.48} & \textbf{24.66} &  & \textbf{89.77} & \textbf{43.92} & \textbf{53.92} \\ \bottomrule
\end{tabular}}
\vspace{-0.1in}
\end{table*}

%% file: tabs/tab2.tex
\centering
\caption{\method's performance improvements on graph oultier detection in AUC, AP, and Recall (\%) on Tolokers.}
\label{tab:boost}
\scalebox{0.75}{
\begin{tabular}{@{}llll@{}}
\toprule
                   & \multicolumn{1}{c}{\textbf{AUC}} & \multicolumn{1}{c}{\textbf{AP}} & \multicolumn{1}{c}{\textbf{Rec}} \\ \midrule
\textbf{GCN \cite{kipf2016semi}} & 75.45 (+0.76)                    & 44.17 (+1.29)                   & 44.24 (+2.18)                    \\
\textbf{SGC \cite{wu2019simplifying}} & 72.73 (+2.06)                    & 39.75 (+1.72)                   & 38.01 (+2.02)                    \\
\textbf{GIN \cite{xu2018powerful}} & 74.83 (+0.78)                    & 38.54 (+1.96)                   & 38.32 (+1.56)                    \\
\textbf{GraphSAGE \cite{hamilton2017inductive}} & 81.65 (+2.23)                    & 52.53 (+3.87)                   & 50.93 (+4.52)                    \\
\textbf{GAT \cite{velivckovic2018graph}} & 82.18 (+4.91)                    & 51.13 (+7.99)                   & 50.78 (+7.48)                    \\
\textbf{GT \cite{shi2021masked}} & 82.73 (+3.49)                    & 51.73 (+5.51)                   & 50.93 (+4.36)                    \\
\hline
\textbf{GAS \cite{li2019spam}} & 76.96 (+0.05)                    & 45.48 (-1.87)                   & 43.15 (-1.87)                    \\
\textbf{DCI \cite{wang2021decoupling}} & 73.75 (-2.23)                    & 37.52 (-2.33)                   & 37.54 (-2.65)                    \\
\textbf{PCGNN \cite{liu2021pick}} & 73.65 (+1.47)                    & 38.42 (+0.90)                   & 38.01 (+1.25)                    \\
\textbf{GATSep \cite{zimek2014ensembles}} & 83.46 (+3.83)                    & 52.95 (+6.87)                   & 52.80 (+6.07)                    \\
\textbf{BernNet \cite{he2021bernnet}} & 76.53 (+0.33)                    & 44.04 (+1.83)                   & 42.21 (+0.00)                    \\
\textbf{AMNet \cite{chai2022can}} & 76.67 (+0.84)                    & 44.31 (+1.65)                   & 43.30 (+1.40)                    \\
\textbf{BWGNN \cite{tang2022rethinking}} & 82.10 (+1.95)                    & 51.94 (+2.29)                   & 51.25 (+3.89)                    \\
\textbf{GHRN \cite{gao2023addressing}} & 82.05 (+2.25)                    & 51.84 (+2.34)                   & 51.56 (+3.27)                    \\ \bottomrule
\end{tabular}}

%% file: tabs/dgraph.tex
\centering
\caption{Performance in AUC, AP, and Rec (\%) on DGraph.}
\label{tab:dgraph}
\scalebox{0.75}{
\begin{tabular}{@{}llll@{}}
\toprule
                     & \textbf{AUC} & \textbf{AP} & \textbf{Rec} \\ \midrule
\textbf{HGT \cite{hu2020heterogeneous}}         & 72.89        & 3.13        & 5.20         \\
\textbf{w/o Feature} & 75.10        & 3.64        & 5.59         \\
\textbf{w/o Edge}    & 74.36        & 3.52	      & 5.76         \\
\textbf{w/o Time}    & 74.23        & 3.29        & 4.73         \\
\textbf{w/o Type}    & 75.09        & 3.62        & 5.98         \\
\textbf{GOVAE}       & 75.36        & 3.58        & 5.54         \\
\textbf{GODM}        & \textbf{75.80}        & \textbf{3.66}        & \textbf{6.19}         \\ \bottomrule
\end{tabular}}

%% file: tabs/data.tex
\begin{table}[ht]
\caption{Statistics of datasets (*: with multiple edge types and temporal infomation).}
\label{tab:data}
\centering
\scalebox{0.8}{
\begin{tabular}[width=\linewidth]{@{}lrrrr@{}}
\toprule
\textbf{Dataset}   & \multicolumn{1}{c}{\textbf{\#Nodes}} & \multicolumn{1}{c}{\textbf{\#Edges}} & \multicolumn{1}{c}{\textbf{\#Features}} & \multicolumn{1}{c}{\textbf{Outlier}} \\ \midrule
\textbf{Weibo}     & 8,405                                & 407,963                              & 400                                    & 10.3\%                               \\
\textbf{Tolokers}  & 11,758                               & 519,000                              & 10                                     & 21.8\%                               \\
\textbf{Questions} & 48,921                               & 153,540                              & 301                                    & 3.0\%                                \\
\textbf{Elliptic}  & 203,769                              & 234,355                              & 166                                    & 9.8\%                                \\
\textbf{DGraph$^*$}    & 3,700,550                            & 4,300,999                            & 17                                     & 1.3\%                                \\ \bottomrule
\end{tabular}}
\end{table}

%% file: tabs/comp.tex
\begin{table}[ht]
\caption{Comparison of data augmentation methods.}
\label{tab:comp}
\centering
\scalebox{0.85}{
\begin{tabular}{@{}lcccc@{}}
\toprule
\textbf{}      & \textbf{Flexibility} & \textbf{Heterogeneous} & \textbf{Time} & \textbf{Diffusion} \\ \midrule
\textbf{DAGAD} & \tickNo         & \tickNo           & \tickNo        & \tickNo         \\
\textbf{GOVAE} & \tickYes        & \tickYes          & \tickYes       & \tickNo         \\
\textbf{GODM}  & \tickYes        & \tickYes          & \tickYes       & \tickYes        \\ \bottomrule
\end{tabular}}
\end{table}

%% file: tabs/api.tex
\begin{minipage}{0.95\linewidth}
\renewcommand{\lstlistingname}{Code Demo} 
\begin{lstlisting}[caption={Using \method on Weibo dataset \cite{zhao2020error}.},captionpos=b, label={lst:label}, numbers=left, xleftmargin=0.5em,frame=single,framexleftmargin=1.3em]
    from pygod.utils import load_data             # import data
    data = load_data('weibo')                     # load weibo data

    from godm import GODM                         # import GODM
    godm = GODM(lr=0.004)                         # initialize GODM             
    aug_data = godm(data)                         # augment data

    detector(aug_data)                            # train on data
\end{lstlisting}
\end{minipage}